\documentclass[sigconf]{acmart}
\usepackage{comment}
\usepackage{graphicx}

\AtBeginDocument{%
  \providecommand\BibTeX{{%
    \normalfont B\kern-0.5em{\scshape i\kern-0.25em b}\kern-0.8em\TeX}}}

\copyrightyear{2021}
\acmYear{2021}
\acmConference[WWW '21 Companion]{Companion Proceedings of the Web Conference 2021}{April 19--23, 2021}{Ljubljana, Slovenia}
\acmBooktitle{Companion Proceedings of the Web Conference 2021 (WWW '21 Companion), April 19--23, 2021, Ljubljana, Slovenia}
\acmPrice{}
\acmDOI{10.1145/3442442.3451382}
\acmISBN{978-1-4503-8313-4/21/04}

\acmSubmissionID{FinW02imS}

\begin{document}

\title{FinMatcher at FinSim-2: Hypernym Detection in the Financial Services Domain using Knowledge Graphs}

\author{Jan Portisch}
\affiliation{%
  \institution{University of Mannheim, Data and Web Science Group}
  \streetaddress{B 6, 26}
  \city{Mannheim}
  \country{Germany}
  \postcode{43017-6221}
}
\email{jan@informatik.uni-mannheim.de}
\orcid{0000-0001-5420-0663}

\author{Michael Hladik}
\affiliation{%
  \institution{SAP SE Product Engineering Financial Services}
  \streetaddress{Dietmar-Hopp-Allee 16}
  \city{Walldorf}
  \country{Germany}}
\email{michael.hladik@sap.com}
\orcid{0000-0002-2204-3138}

\author{Heiko Paulheim}
\orcid{0000-0001-5420-0663}
\affiliation{%
  \institution{University of Mannheim, Data and Web Science Group}
  \streetaddress{ B 6, 26}
  \city{Mannheim}
  \country{Germany}
  \postcode{43017-6221}
}
\email{heiko@informatik.uni-mannheim.de}
\orcid{0000-0003-4386-8195}

\renewcommand{\shortauthors}{Portisch et al.}

\begin{abstract}
  This paper presents the \emph{FinMatcher} system and its results for the FinSim 2021 shared task which is co-located with the Workshop on Financial Technology on the Web (FinWeb) in conjunction with The Web Conference. The FinSim-2 shared task consists of a set of concept labels from the financial services domain. The goal is to find the most relevant top-level concept from a given set of concepts.
  The FinMatcher system exploits three publicly available knowledge graphs, namely WordNet, Wikidata, and WebIsALOD. The graphs are used to generate explicit features as well as latent features which are fed into a neural classifier to predict the closest hypernym.
\end{abstract}



\maketitle

\section{Introduction}
A \emph{hypernym} or \emph{hyperonym} is a concept which is superordinate to another one. In computer science, it is often represented as an \emph{IS-A} relationship. For example, \emph{animal} is a hypernym of \emph{cat} and \emph{equity index} is a hypernym of \emph{S\&P 500 Index}. A \emph{hyponym}, on the other hand, is a concept which is subordinate to another one. For example, \emph{cat} is a hyponym of \emph{animal} and \emph{S\&P 500 Index} is a hyponym of \emph{equity index}.~\cite{murphy2003semantic} Hypernymy detection can be broadly applied in real-world applications. The detection of hypernyms in the financial services domain is particularly interesting due to a domain specific vocabulary and a lack of publicly available domain-specific resources and concept representations. 

The FinSim task models the hypernym detection task as a multi class classification problem: Given a concept label (i.e., the hyponym), the correct hypernym is to be found from a set of 10 mutually exclusive classes (i.e., hypernyms). A system participating in this task can return a sorted list of classes. The task is evaluated with two performance metrics: mean rank and accuracy. 

The FinMatcher system uses two very broad publicly available knowledge graphs (Wikidata and WebIsALOD) as well as a small linguistic graph resource (WordNet). 
A knowledge graph contains real world entities from various domains and the relationships that hold between them in a graph format~\cite{DBLP:journals/semweb/Paulheim17}. 
The system presented in this paper calculates multiple explicit features and uses RDF2vec embeddings obtained from WebIsALOD. The features are concatenated into a feature vector which is presented to a neural classifier which was trained with the provided FinSim training data. 

In the following section, related work is introduced. Afterwards, the provided dataset is quickly described. 
In Section~\ref{sec:system_description}, the FinMatcher system is presented.
The results of FinSim task are given in Section~\ref{sec:results} together with an ablation study. The paper is concluded in Section~\ref{sec:conclusion} where future research directions are also presented.

\section{Related Work}

\subsection{Shared Tasks for Hypernym Detection}
Hypernym discovery has been addressed before as challenge, for example at SemEval-2018~\cite{DBLP:conf/semeval/Camacho-Collados18}. Unique to the FinSim task is the focus on the financial services industry. The evaluation campaign premiered in 2020~\cite{el2021finsim} and has been extended for the 2021 campaign, also referred to as FinSim-2~\cite{finsim_2021}: Two additional tags have been introduced and the training and evaluation datasets have been extended.

\subsection{Knowledge Graphs}
FinMatcher uses three external knowledge graphs as background knowledge for the task of hypernym detection. 

\emph{WordNet}~\cite{wordnet} is a well known lexical resource. It is a database of English words grouped in sets which represent a particular meaning, called \emph{synsets}; further semantic relations such as hypernymy also exist in the database. The resource is publicly available.\footnote{see \url{https://wordnet.princeton.edu/download}} 

\emph{Wikidata} is a knowledge graph hosted by the Wikimedia Foundation which is publicly available\footnote{see \url{https://www.wikidata.org/wiki/Wikidata:Main_Page}} and maintained by an open community. The graph contains class-like entities, such as ``stock market index'', and also instance-like entities, such as ``MSCI World''. An example for a Wikidata statement would be \emph{``MSCI World'' instance of ``stock market index''}\footnote{see \url{https://www.wikidata.org/wiki/Q1881843}}. The graph can be queried using SPARQL\footnote{see \url{https://query.wikidata.org/}}.

A frequent problem that occurs when working with external background knowledge in the financial services domain is the fact that less common entities -- so called \emph{long tail entities} -- are not contained within a knowledge base. The \emph{WebIsA}~\cite{seitner_large_2016} database is an attempt to tackle this problem by providing a dataset which is not based on a single source of knowledge -- like \emph{DBpedia}~\cite{lehmann_dbpedia_2012} -- but instead on the whole Web: The dataset consists of hypernymy relations extracted from the \emph{Common Crawl}\footnote{see \url{http://commoncrawl.org/}}, a freely downloadable crawl of a significant portion of the Web. For the automated extraction, lexico-syntactic patterns similar to those presented by Hearst~\cite{DBLP:conf/coling/Hearst92} were used.
Like Wikidata, the graph contains class-like and instance-like concepts. A sample triple from the dataset is \textit{``zero-coupon bond'' skos:broader ``bond''}\footnote{see \url{http://webisa.webdatacommons.org/concept/zero-coupon_bond_}}. 
The dataset is also available via a Linked Open Data (LOD) endpoint\footnote{see \url{http://webisa.webdatacommons.org/}} under the name \emph{WebIsALOD}~\cite{hertling_webisalod:_2017} -- hence, it can be queried like Wikidata using SPARQL.

\subsection{Knowledge Graph Embeddings}
In recent years, latent representations have gained traction not only in natural language processing but also in other data science communities. \emph{RDF2vec}~\cite{rdf2vec} is a knowledge graph embedding approach which allows to obtain a latent representation for the elements of a knowledge graph, i.e. a vector, for each node and each edge in a graph. It applies the \emph{word2vec}~\cite{word2vec_1,word2vec_2} model to RDF data: Random walks are performed for each node and are interpreted as sentences. After the walk generation, the sentences are used as input for the word2vec algorithm. As a result, one obtains a vector for each word, i.e., a concept in the RDF graph. Multiple flavors of RDF2vec have been developed in the past such as biased walks~\cite{DBLP:conf/wims/CochezRPP17} or \emph{RDF2vec Light}~\cite{DBLP:conf/semweb/PortischHP20}.\footnote{For a good overview of the RDF2vec approach and its applications, refer to\\ \url{http://www.rdf2vec.org/}} The calculation of knowledge graph embeddings on large graphs can require a significant amount of resources. Therefore, \emph{KGvec2go}\footnote{see \url{http://www.kgvec2go.org}}~\cite{DBLP:conf/lrec/PortischHP20} provides pre-trained RDF2vec knowledge graph embeddings through a Web API as well as via download. For the system presented in this paper, a pre-trained embedding of WebIsALOD has been downloaded from KGvec2go. 

Both, RDF2vec and WebIsALOD, have been used for integration tasks in the financial services domain before~\cite{DBLP:conf/semweb/MonychPHP20,portisch2018automatic}. 

\section{FinSim Dataset Description}
\begin{figure*}
    \centering
    \includegraphics[scale=0.3]{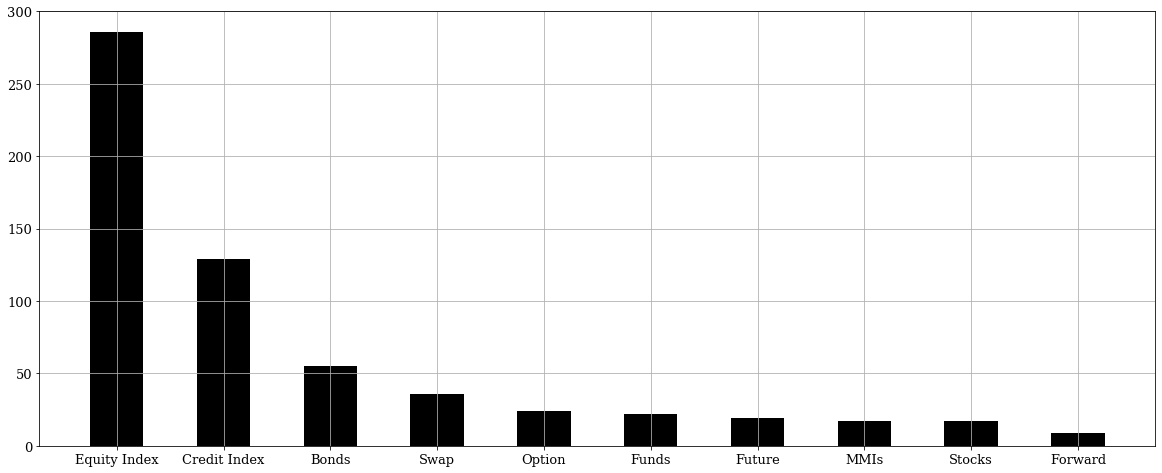}
    \caption{Distribution of class labels in the FinSim training dataset}
    \label{fig:class_label_distribution}
\end{figure*}

The FinSim dataset consists of 614 hyponym-hypernym pairs. There are 10 class labels, i.e. hypernyms: 
\begin{enumerate}
    \item Equity Index
    \item Credit Index
    \item Bonds
    \item Swap
    \item Option
    \item Funds
    \item Future
    \item MMIs
    \item Stocks
    \item Forward
\end{enumerate}
The class labels presented above classify concepts not according to their features but instead according to their prototypical kind. 
The distribution of class labels is not balanced. As shown in Figure~\ref{fig:class_label_distribution}, the distribution of labels follows a power-law with 286 entries for ``equity index'' and only 9 entries for ``forward''. This is a challenging setting for multiple reasons: (i) the training dataset is comparatively small, (ii) the hypernyms are semantically very related, (iii) industry abbreviations are used, and (iv) there are textual overlaps. The FinSim-2 test dataset consists of 212 entries, the distribution of class labels is not known. 

Compared to other evaluation campaigns where participants have to submit their implementations, such as the Ontology Alignment Evaluation Initiative (OAEI), participants of the FinSim task run their system on their own premises and submit the predictions made by their system.

\section{System Description}
\label{sec:system_description}
The FinMatcher system combines explicit and latent features. In total, there are five groups of features which will be presented in the following. The overall architecture is shown in Figure~\ref{fig:architecture}.

\subsection{Features}
\begin{figure*}
    \centering
    \includegraphics[scale=0.3]{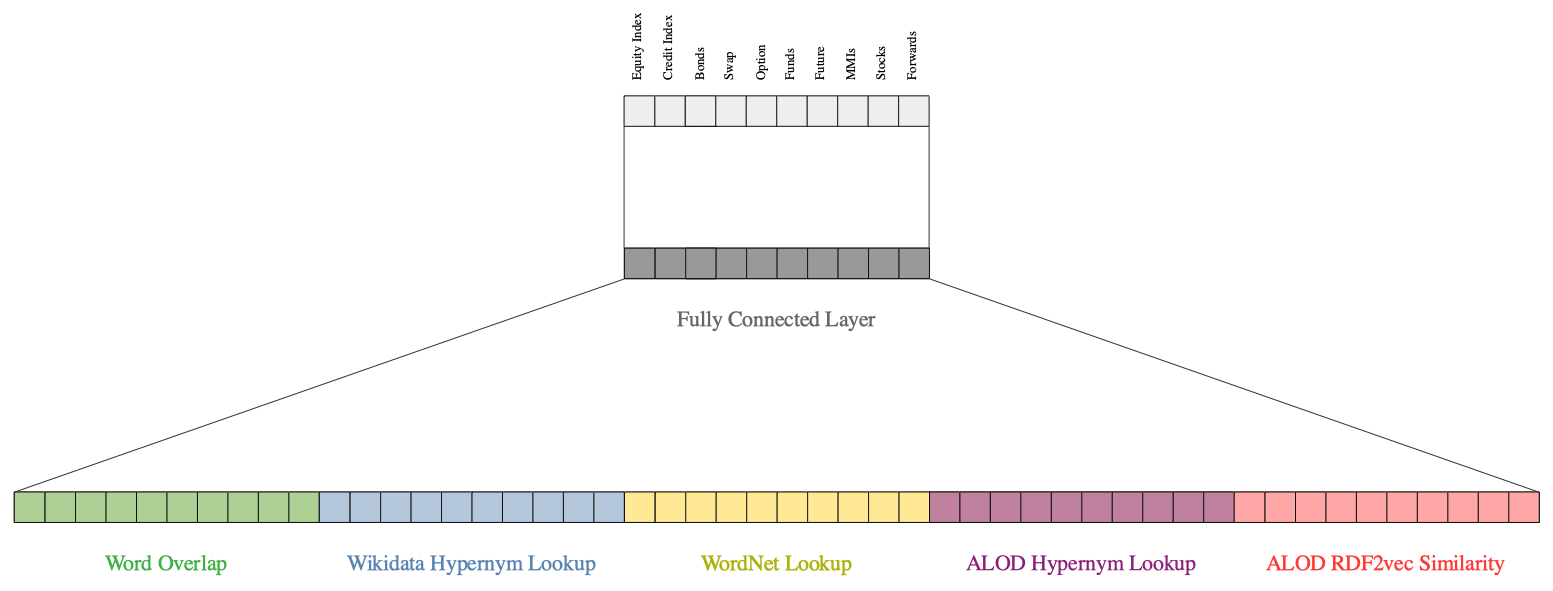}
    \caption{Architecture of FinMatcher}
    \label{fig:architecture}
\end{figure*}

\paragraph{Word Overlap} The overlap between hyponym and class label is a strong signal for a match. An example would be ``Supranational Bond'' which is a ``Bond''. As such constellations are relatively frequent in the provided dataset, the first feature vector encodes whether the label contains the class label. For this feature minimal text pre-processing is applied including lower-casing and removal of the plural suffix ``s''. As this step is performed for each class label, a vector of length 10 is obtained. The overlap feature vector is displayed in green in Figure~\ref{fig:architecture}.

\paragraph{Wikidata Hypernym Lookup} Wikidata is a large general-purpose knowledge graph which is not tailored to the financial domain. Nonetheless, the data source contains many financial concepts and relations between them. For example, the concept ``UCITS'' can be linked to``Undertakings for Collective Investment in Transferable Securities'' via the \emph{also known as} label; due to the annotated relation \emph{subclass of}, it is easily recognizable that ``UCITS'' is an ``investment fund''.\footnote{see \url{https://www.wikidata.org/wiki/Q25323628}} This notion is exploited in this set of features: A comprehensive linking mechanism from the MELT framework\footnote{The Matching EvaLuation Toolkit (MELT) is a framework for ontology and instance matching (development, evaluation, visualization~\cite{DBLP:conf/esws/PortischHP20}). However, components can also be exploited for other tasks. For a better overview, see \url{https://github.com/dwslab/melt/}}~\cite{DBLP:conf/i-semantics/HertlingPP19,DBLP:conf/semweb/HertlingPP20} is used to link classes (the hypernyms) as well as labels (the hyponyms) to Wikidata concepts and then relations $P31$ (instance of) and $P279$ (sublcass of) are followed up to two hops to evaluate whether the class label appears. Distant matches receive a lower signal strength which is calculated through the inverse hop-distance: A direct hypernym annotation (as in the UCITS example stated earlier) receives the value $\frac{1}{1} = 1$ whereas a two-hop match would receive a value of $\frac{1}{2} = 0.5$. As this step is performed for each class label, a vector of length 10 is obtained. The Wikidata lookup feature vector is displayed in blue in Figure~\ref{fig:architecture}.

\paragraph{WordNet Hypernym Lookup} The same exploitation approach chosen for Wikidata is applied on the WordNet graph: Hypernyms and hyponyms are linked into WordNet and then the inverse hop-distance is used as feature value. This is done for each class label that could be linked. The WordNet lookup feature vector is displayed in yellow in Figure~\ref{fig:architecture}.

\paragraph{WebIsALOD Hypernym Lookup} In a similar fashion to the Wikidata hypernym lookup, class labels as well as hyponym labels are linked to the WebIsALOD graph using a linker from the MELT framework. In this graph, there exists only one significant relation: \emph{skos:broader}. For each hyponym, the broader concepts are obtained and it is checked whether the hypernym appears. Due to a high level of noise, the number of upwards hops is limited to 1. As this step is performed for each class label, a vector of length 10 is obtained. The WebIsALOD lookup feature vector is displayed in purple in Figure~\ref{fig:architecture}.

\paragraph{WebIsALOD RDF2vec Similarity} For the embedding feature, each class label as well as each hyponym label is linked again into the WebIsALOD knowledge graph. Each concept in Web\-IsALOD has an associated embedding vector $v \in I\!R^{200}$. For comparisons, the cosine similarity between the hyponym and the class label is calculated.

If the whole concept cannot be linked, multiple sub-concepts are detected and linked. Within this linking process, longer sub-concepts are favored. For example, the string ``CDX Emerging Markets'' cannot be directly linked -- however, the longest substring that can be linked here is ``Emerging Markets''; in addition, ``CDX'' can also be linked. Comparisons in such cases are performed as follows:

\begin{equation}
    \frac{\sum^I_{i=0} \max^J_{j=0}(sim(v_i, v_j)}{|I|}
\end{equation}
where $I$ represents the set of links of the hyponym, $J$ represents the set of links of the hypernym, $v_i$ and $v_j$ correspond to the vectors of the links and $sim$ refers to a similarity function. In this case, the cosine is used as similarity function.
As this step is performed for each class label, a vector of length 10 is obtained. The WebIsALOD lookup feature vector is displayed in salmon in Figure~\ref{fig:architecture}.

\paragraph{Feature Composition} Each of the features $i$ returns a signal vector $s_i \in I\!R^{10}$. All vectors are concatenated to form the final signal vector  $S = \mathbin\Vert ^{5}_{i=1}s_i$, which is used as input for the classifier. 

\subsection{Classifier}
Due to the small total number of training examples, a very simple artificial neural network architecture has been chosen. It is configured with one fully connected layer of size 10 and mean squared error as loss. The network was trained with 100 epochs and a batch size of 25 on a consumer PC. The vector that is to be predicted is of size 10 and represents the one-hot-encoded class label. The neural network classifier performed best among the classifiers evaluated: Na\"{i}ve Bayes, J48 decision trees, random forests, and a regression. 

As the distribution of class labels is skewed (see Figure~\ref{fig:class_label_distribution}), we applied the synthetic minority oversampling technique (SMOTE)~\cite{smote} to upsample underrepresented class labels. We experimentally chose 33\% of the majority class total as the upsampling barrier; this means that if the majority class in the training split totals to 229 records, upsampling for class labels with less than $\frac{1}{3}*229 = 76$ records will be performed so that there are 76 records for the underrepresented class label. 

\section{Results}
\label{sec:results}

\subsection{Results on the Training Data and Ablation Study}
We evaluated our matching system by performing a stratified five-fold cross validation on the training data. We trained each ANN configuration 10 times and report the average results for accuracy and mean rank. We further performed an ablation study by training and evaluating the performance when leaving out each of the five feature groups. The results can be found in Table~\ref{tab:ablation}. 

It is visible that the most important feature group in terms of accuracy is word overlap. This is not surprising given the high number of labels that contain the hypernym within their name (for example ``green bonds'' $\rightarrow$ ``bonds'') and shows that it is sensible for the task at hand to combine explicit and latent features. The observation that the inclusion of the target label in the term is a significant signal has also been made in the last FinSim campaign~\cite{el2021finsim}.  
The negligible role of WordNet in terms of accuracy is also comprehensible since this particular external background knowledge dataset contains merely general-purpose class knowledge (such as ``call option'') but no knowledge about instances (such as ``MSCI EMU Index''). For the FinSim dataset, very large knowledge graphs that contain class as well as instance knowledge are more beneficial due to their higher concept coverage. However, the information in the knowledge graphs used also contain some redundancy, as can be observed in Table~\ref{tab:ablation}: leaving out a single knowledge graph does not significantly change the results.

\begin{table}
  \caption{Ablation Study}
  \label{tab:ablation}
  \begin{tabular}{ccc}
    \toprule
    Left-out Vector Group & Mean Accuracy & Mean Rank (HITS@10)\\
    \midrule
    Submission & 86.69 & 1.432  \\
    No SMOTE & 85.55 & 1.371 \\ 
    Word Overlap & 60.51 & 2.007\\
    Wikidata Hypernyms & 85.50 & 1.490 \\
    WordNet Hypernyms & 86.66 & 1.440 \\
    ALOD Hypernyms & 85.88 & 1.361 \\
    ALOD RDF2vec & 85.56 & 1.481 \\
  \bottomrule
\end{tabular}
\end{table}

To further analyze the contribution of the different signals, we plotted the weights of the input features. As the weight of each input neuron $s_i$ relates to label $i$, we can directly observe which features the trained model considers relevant to identify which label.

\begin{figure*}
    \centering
    \includegraphics[scale=0.25]{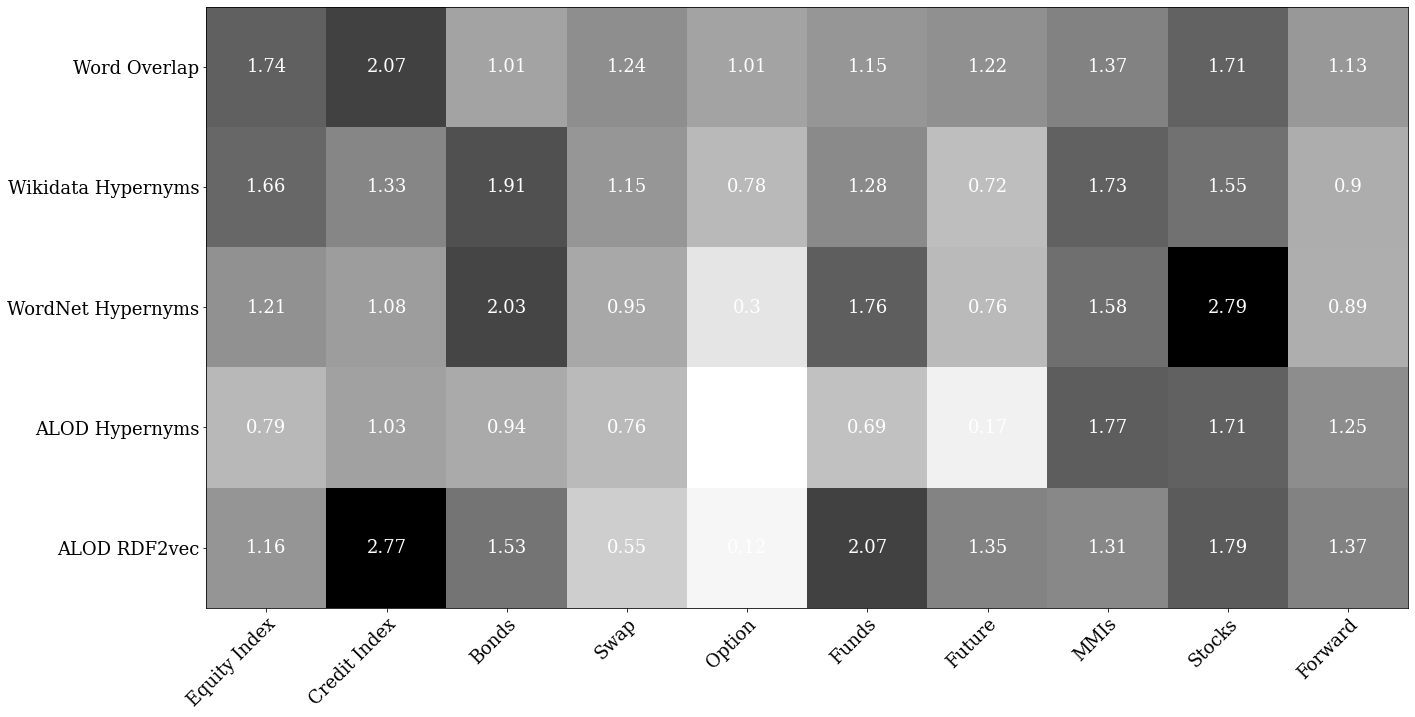}
    \caption{Heatmap of the Absolute Weights per Feature Group and Class Label}
    \label{fig:weights_heatmap}
\end{figure*}

Table~\ref{tab:weights} shows the summed absolute weight per feature group. This allows to analyze the overall contribution of the individual feature group. Here, it is visible that the latent RDF2vec feature group receives the highest weight -- higher than the word overlap group. 
While the word overlap feature is important for the majority labels (equity index, credit index), it is not equally important for all labels and does not have the overall highest weight: Figure~\ref{fig:weights_heatmap} shows the summed absolute weight per feature group and class label. The class labels are sorted in descending order by frequency. Here, it is visible that the word overlap has the highest contribution for the equity index as well as a high contribution for the credit index but low weights for the remaining minority classes. 

\begin{table}[H]
    \caption{Absolute Weights per Feature Group}
    \label{tab:weights}
    \begin{tabular}{cc}
    \toprule
    Vector Group & Weight\\
    \midrule
    Word Overlap & 13.64 \\
    Wikidata Hypernyms & 13.01 \\
    WordNet Hypernyms & 13.35 \\
    ALOD Hypernyms & 9.12 \\
    ALOD RDF2vec & 14.01\\
    \bottomrule
    \end{tabular}
\end{table}

\subsection{Results on the Reference Data}
FinMatcher participated only with one configuration and achieved an accuracy of 81.1\% and a mean rank of 1.415 on the reference data below the expected scores from the training data shown in Table~\ref{tab:ablation}.

\section{Conclusion}
\label{sec:conclusion}
In this paper, we presented \emph{FinMatcher}, a hypernym detection system for the financial services domain which exploits multiple knowledge graphs by combining explicit and latent features. We could show that the task can be addressed by including external knowledge in the form of knowledge graphs and that the combination of multiple graphs is overall beneficial.

In the future, we strive to improve the results through the inclusion of more advanced embedding techniques as well as the exploration of additional external datasets. 

\begin{acks}
We would like to thank the FinSim-2 organizers (Youness Mansar, Isma\"{i}l El Maarouf, and Juyeon Kang) for compiling the data, conducting the evaluation campaign, and for promptly answering all questions.
\end{acks}

\bibliographystyle{ACM-Reference-Format}
\bibliography{references}

\end{document}